\documentclass[11pt]{article}
\usepackage{acl2017}
\usepackage{times}
\usepackage{url}
\usepackage{latexsym}
\usepackage{xspace}
\usepackage{booktabs}
\usepackage{color}
\usepackage{graphicx}
\usepackage{tabulary}
\usepackage{enumitem}
\usepackage{multirow}
\usepackage{amsfonts}

\newenvironment{tightitemize}%
  {\begin{itemize}[topsep=0pt, partopsep=0pt] %
    \setlength{\itemsep}{0pt}%
    \setlength{\parskip}{0pt}%
    }%
  {\end{itemize}}
  \usepackage{subcaption}
\usepackage{lipsum}
\usepackage{multirow}

  \usepackage{color}
  {\begin{enumerate}≈ \footnotesize%
    \setlength{\itemsep}{0pt}%
    \setlength{\parskip}{0pt}%
    }%
  {\end{enumerate}}

\usepackage{microtype}
\usepackage{multirow}
\usepackage{verbatim}
\usepackage{amsmath,amsthm,amssymb}
\usepackage{array}
\usepackage[scaled=0.86]{helvet}
\usepackage{ifthen}
\usepackage{courier}
\usepackage[linesnumbered,vlined,ruled]{algorithm2e}
\newcommand{\sysname}{\textsc{IcoRating}\xspace}
\newcommand{\captionfonts}{\small}
\makeatletter  
\long\def\@makecaption#1#2{%
  \vskip\abovecaptionskip
  \sbox\@tempboxa{{\captionfonts #1: #2}}%
  \ifdim \wd\@tempboxa >\hsize
    {\captionfonts #1: #2\par}
  \else
    \hbox to\hsize{\hfil\box\@tempboxa\hfil}%
  \fi
  \vskip\belowcaptionskip}
\makeatother   

\belowdisplayskip \abovedisplayskip

\aclfinalcopy

\title{IcoRating:  A Deep-Learning  System for Scam ICO Identification}
\author{Shuqing Bian$^1$,
Zhenpeng Deng$^1$,
  {\bf  Fei Li}$^3$,
 Will Monroe$^4$, 
Peng Shi$^1$,
Zijun Sun$^1$, \\
 {\bf Wei Wu}$^{1}$,
  {\bf Sikuang Wang}$^1$, 
 {\bf William Yang Wang}$^2$, 
   {\bf Arianna Yuan}$^{1~4}$, \\  
     {\bf Tianwei Zhang}$^1$
  and  {\bf Jiwei Li}$^{1~4}$ ~~\\~~\\
{$^1$Shannon.AI} \\
{$^2$University of California, Santa Barbara } \\
{$^3$University of Michigan, Ann Arbor } \\
{$^4$Stanford University } \\ 
}
\date{}
\aclfinalcopy
\begin{document}
\maketitle

\begin{abstract}
Cryptocurrencies (or digital tokens, digital currencies, e.g., BTC, ETH, XRP, NEO) 
have 
been rapidly gaining ground in use, value, and understanding among the public, 
bringing 
astonishing profits to investors.
Unlike other money and banking systems, 
most digital tokens do not require central authorities.
Being decentralized  poses significant challenges for credit rating. 
Most ICOs are currently not subject to government regulations, which makes
 a reliable  credit rating system for ICO projects necessary and urgent. 

In this paper, we introduce \sysname,
the first  learning--based cryptocurrency rating system. We exploit  natural-language processing techniques to analyze various aspects of 2,251 digital currencies to date, such as white paper content, founding teams, Github repositories, websites, etc. Supervised learning models are used to  correlate the life span and the price change of cryptocurrencies with these features. For the best setting, the proposed system is able to identify scam ICO projects with 0.83 precision.

We hope this work will help investors identify scam ICOs
and attract more efforts in automatically evaluating and analyzing ICO projects.
  \footnote{Author contributions: J.~Li designed research; 
Z.~Sun, Z.~Deng, F.~Li and P.~Shi
 prepared the data; 
S.~Bian and A.~Yuan contributed  analytic tools; 
P.~Shi and Z.~Deng labeled the dataset;
J.~Li, W.~Monroe and W.~Wang designed the experiments;
J.~Li, W.~Wu, Z.~Deng and T.~Zhang performed the experiments;
J.~Li and T.~Zhang wrote the paper;  W.~Monroe and A.~Yuan proofread the paper.
} 
\footnote{Author Contacts: \\
Jiwei Li (corresponding author):
jiwei$\_$li@shannonai.com;  \\
William Yang Wang: william@cs.ucsb.edu \\
Will Monroe:  wmonroe4@stanford.edu\\
Arianna Yuan: xfyuan@stanford.edu
}

\end{abstract}
\begin{figure}[!ht]
    \centering
        \includegraphics[width=3in]{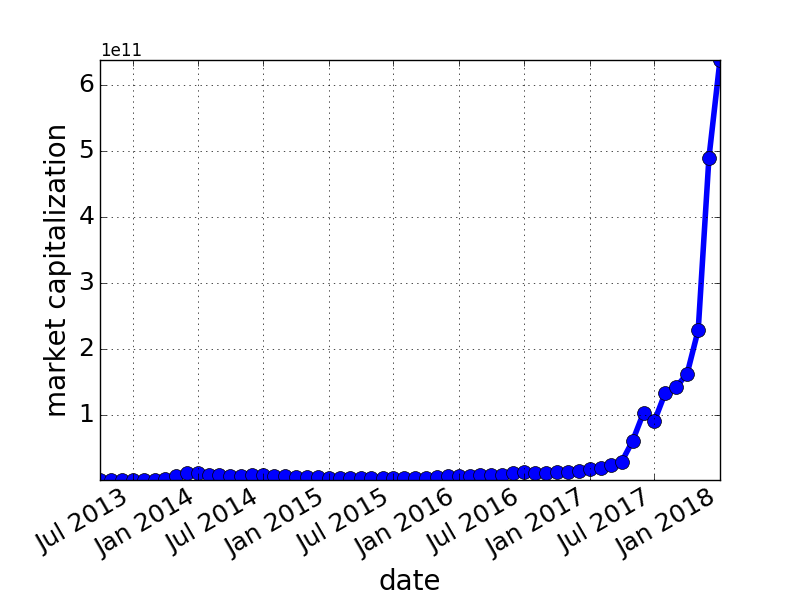}
        \caption{Market capitalization v.s. time.}
        \label{cap-generative}
\end{figure}
\begin{figure}[!ht]
    \centering
        \includegraphics[width=3in]{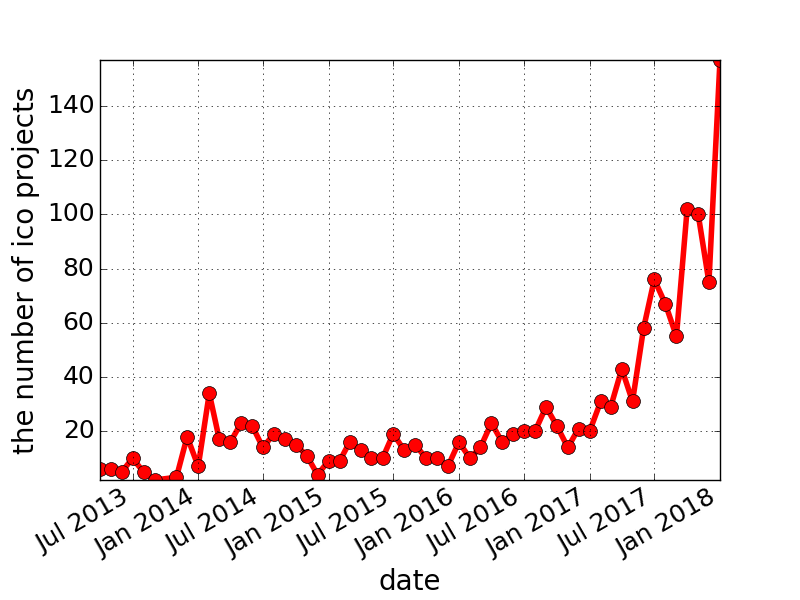}
        \caption{The number of new ICO projects v.s. time.}
        \label{ico-generative}
\end{figure}

\section{Introduction}
Cryptocurrencies (e..g, BTC, ETH, NEO) are gaining unprecedented popularity and  understanding. 
As opposed to centralized electronic money and central banking systems, 
most digital tokens do not require central authorities. 
The control of these 
decentralized systems
 works through a blockchain, which is an open and distributed ledger that continuously grows.
 The market capitalization of cryptocurrencies has increased by a significant margin over the past 3 years, as can be seen from Figure \ref{cap-generative}. 
   According to data provider 
 Cryptocurrency Market Capitalizations,\footnote{https://coinmarketcap.com/},
      the peak 
daily trading volume of cryptocurrencies
is 
close to the average daily volume of trade on the New York Stock Exchange in 2017.

Because of their decentralized nature, 
the crowdfunding of 
digital coins 
 does not need to go through all the required conditions of Venture Capital investment, but through initial Coin Offerings (ICOs) \cite{Ch:17}.
In an ICO,  investors obtain the crowdfunded cryptocurrency using legal tender (e.g., USD, RMB) or  other cryptocurrencies (e.g., BTC, ETH), and these 
crowdfunded cryptocurrencies become functional units of currency when the ICO is done. 
A white paper is often prepared prior to launching the new cryptocurrency, detailing the commercial, technological and financial details of the coin.  
As can be seen in Figure \ref{ico-generative}, the number of ICO projects grew steadily from July 2013 to January 2017, and skyrocketed in 2017.

\begin{figure}
    \centering
        \includegraphics[width=3.2in]{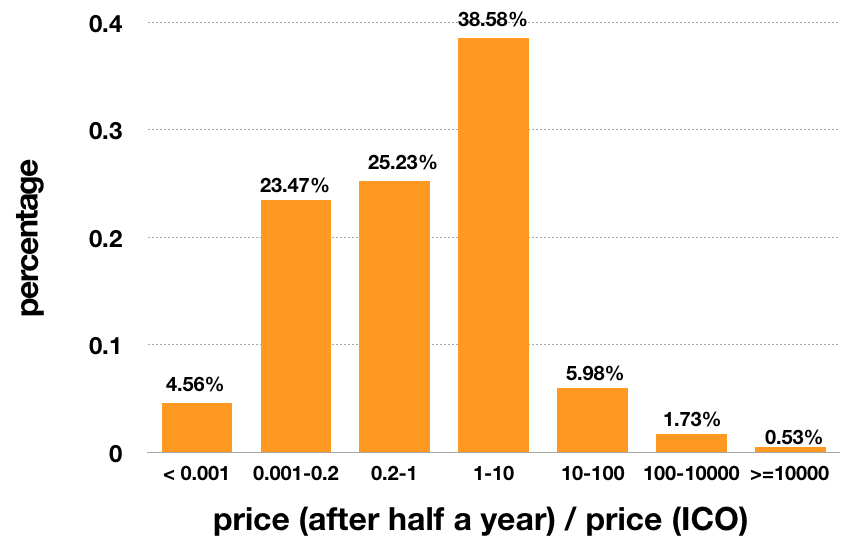}
        \caption{ ICO Percentage vs  half-year price change.}
        \label{half-year}
\end{figure}

\begin{figure}
    \centering
        \includegraphics[width=3.2in]{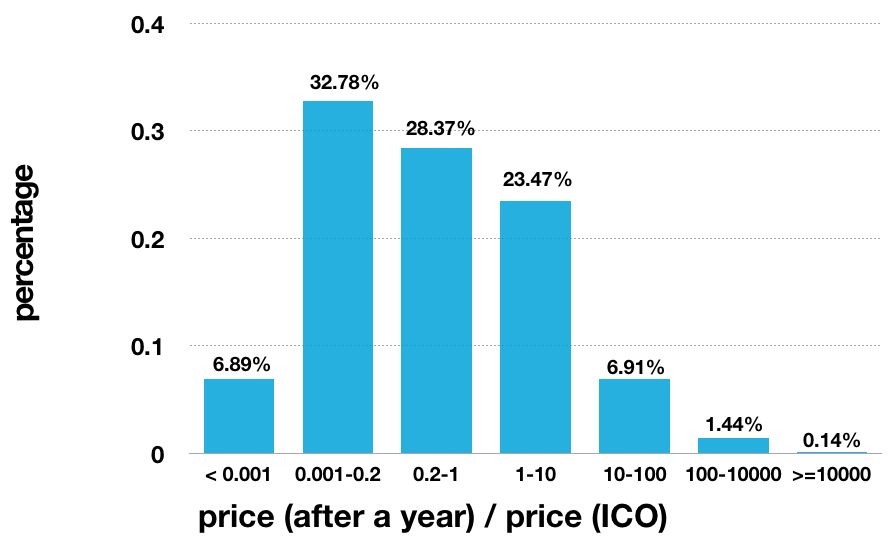}
        \caption{ICO Percentage vs one-year price change.}
        \label{one-year}
\end{figure}

Despite the fact that 
ICOs are able to 
provide fair and lawful investment opportunities, 
the ease of crowdfunding 
creates opportunities and incentives
for unscrupulous businesses to use 
 ICOs to execute ``pump and dump'' schemes, in which the ICO initiators drive up the value of the crowdfunded cryptocurrency and then quickly ``dump'' the coins for a profit.  
Additionally, the  decentralized nature of cryptocurrencies
 poses significant challenges for government regulation. 
 According to Engadget,\footnote{\url{https://www.engadget.com/}}
  46 percent of the 902 crowdsale-based digital currencies
  in 2017
   have already failed. 
Figures \ref{half-year} and \ref{one-year} show an even more serious situation.
Each bucket on the x axis 
of Figures \ref{half-year} and \ref{one-year}
denotes a range of price change, and the corresponding value on the y axis denotes the percentage of ICO projects.
As can be seen,
$4.56\%$ of existing ICO projects suffered a price fall of more than $99.9\%$ after half a year, 
and this fraction goes up to $6.89\%$  after one year. 
About $29\%$ of projects fell more than $80\%$ after half a year, increasing to
an astonishing value of
$39.6\%$  after a year.
Though it is not responsible to say that every sharply-falling ICO project is a scam, 
it is necessary and urgent to build a reliable ICO credit rating system to evaluate a digital currency before its ICO. 

In this paper, we propose
\sysname,
a machine-learning based ICO rating system. 
By analyzing 2,251 ICO projects,
we correlate
the life span and the price change  of a digital currency with various levels of its ICO information, including
 its white papers, founding team, GitHub repository, website, etc. 
For the best setting, the proposed system is able to identify scam ICO projects with a precision of 0.83  and an F1 score of 0.80.

\sysname
is a machine learning--based system. 
Compared against human-designed rating systems, \sysname has two key advantages.
(1) {\bf Objectivity}: 
 a machine learning model
involves less prior knowledge about the world, instead learning the causality 
from the data, in contrast to
 human-designed systems
that require
massive
 involvement of human experts, who inevitably introduce biases. 
(2) {\bf Difficulty of manipulation} by unscrupulous actors: 
 the credit rating result is output from a machine learning model through black-box training. 
 This process requires minor human involvement and intervention. 

\sysname can contribute to the cryptocurrency community in two aspects. First, we hope that this work would encourage more efforts invested in designing reliable, automatic and difficult-to-be-manipulated systems to analyze and evaluate the quality of ICO projects. Second, \sysname can potentially help investors identify scam ICO projects and make rational investment in cryptocurrency.

The rest of this paper is organized as follows: we give a brief outline of cryptocurrencies, blockchains and ICOs in Section 2. 
We describe the
construction of a dataset of ICO projects in Section 3, 
and provide some basic analysis of the data.
 In Section 4, we describe the machine learning model we propose。
 Experimental results
 and qualitative analysis
  are illustrated in Section 5,  
  followed by a short conclusion in Section 6. 

\section{Cryptocurrencies, Blockchains, and ICOs}
In this section, we briefly describe relevant information on cryptocurrencies, blockchains, and ICOs. 

\paragraph{Cryptocurrencies}
A cryptocurrency is ``a digital asset designed to work as a medium of exchange that uses cryptography to secure its transactions''.\footnote{Quoted from \url{https://en.wikipedia.org/wiki/Cryptocurrency}.}
Most 
cryptocurrencies use decentralized control.
The first decentralized cryptocurrency is Bitcoin (BTC for short) \cite{nakamoto2008bitcoin}, created in 2009 by  an unknown person or group of people  under the name Satoshi Nakamoto. 
Since the design of BTC, various types of cryptocurrencies have been created,  the most well-known of which include Ethereum \cite{Co:16},
Ripple \cite{Ca:14},
EOS \cite{Co:17} and
NEO \cite{St:17}.

\paragraph{Blockchains}
A cryptocurrency's transactions  are validated by  a blockchain.
One can think a blockchain as a distributed ledger, which 
continuously grows and 
records
all
 transactions between two parties permanently.
Each record is called a block, which contains a cryptographic hash pointer that links to a previous block, a timestamp and transaction data.
 The ledger is owned  in a distributed way by all participants,
and the record cannot be altered  without the alteration of all subsequent blocks  of the network.
Transactions are broadcast to all nodes in the network. Blockchains use various time-stamping schemes such as proof-of-work \cite{DwNa:93} or proof-of-stake \cite{Va}.

The concept of blockchain eliminates the risks of data being held centrally: it has no central point of failure and data is transparent to every participant involved.

\paragraph{ICOs}
An initial coin offering (ICO)\footnote{\url{https://en.wikipedia.org/wiki/Initial_coin_offering}}
is a means of crowdfunding centered around cryptocurrency. 
In an ICO, 
crowdfunded cryptocurrency (mostly in the form of tokens) is transferred to investors 
in exchange for legal tender or other cryptocurrencies. 
These tokens become functional units of currency that can be exchanged for goods or other cryptocurrencies when the ICO's funding goal is met. 
 
ICOs provide a crowdfunding opportunity for early-stage projects 
 to avoid  
the  regulations required by venture capitalists, bank and stock exchanges. 
They also provide investment opportunities
beyond venture capital or private equity investments, which have dominated early-stage investment opportunities. 
On the other hand, because of the lack of regulations, ICOs pose significant risks for investors. 
Different countries have different regulations on ICOs and  cryptocurrencies.
For example, the government of the People's Republic of China banned all ICOs,  
and the U.S.\ Securities and Exchange Commission (SEC) indicated that it could have the authority to apply federal securities law to ICOs \cite{Stan:17},
while 
the government of Venezuela launched its own cryptocurrency 
called {\it petromoneda}.

\section{Datasets Construction, Processing and Analysis}
We collected information for 2,251 past ICO projects, including the white papers, website information, GitHub repositories at the time of ICO, and founding teams. We obtained the data from various providers including 
CryptoCompare,\footnote{\url{http://cryptocompare.com/}}
CoinMarketCap\footnote{\url{https://coinmarketcap.com/}} and
CoinCheckup.\footnote{\url{http://coincheckup.com/}}

\subsection{ Analysis on White Papers}
The white paper is a crucial part of an ICO project. It describes how the crowdfunding is intended to work, such as 
the landscape of the ICO project, 
how the tokens will be allocated or how the crowdfunded money will be spent. 
Out of the 2,251 ICO projects, we were able to obtain 1,317 white papers.  
We transform white paper PDFs to texts using the PDFMiner API.\footnote{\url{https://github.com/euske/pdfminer}}
\begin{table}[ht]
\center
\small
\begin{tabular}{ccccc}\hline
$\#$Doc& Ave Word &Std Word & Max Word & Min Word \\
1,317 & 7,834 & 7,060 & 13,721&685 \\\hline
$\#$Doc& Ave Sent &Std Sent & Max Sent & Min Sent \\
1,317 & 953 & 1,434 &6,228 &38 \\\hline
\end{tabular}
\caption{Statistics for publicly available ICO projects: average/min/max/standard deviation for number of words and number of sentences
in a white paper. 
}
\label{Statistics}
\end{table}

We show the statistics for ICO white papers in Table \ref{Statistics}, including
average, standard deviation, maximum and minimum numbers of words and sentences in the white paper.
A conspicuous feature of Table \ref{Statistics} is the high variance in the length of white papers, with a maximum of 6,228 sentences and a minimum of 38. 
More concretely, the number of sentences in 10 randomly sampled white papers is 
 [886, 143, 38, 967, 3379, 6228, 496, 2057, 3075, 298]. 
Though the length of a white paper does not necessarily reflect the quality of an ICO project, we can see the large variance in content among ICO white papers.

\begin{figure}
\begin{tabular}{l}\hline\hline
1. sample $\theta^m\sim Dir(\alpha)$\\
2. sample $\phi^k\sim Dir(\beta)$\\
3. for each white paper $m\in [1,M]$: \\
~~~~~~(i) sample topic $z\sim multinomial(\alpha)$\\
~~~~~~(ii) for each word $w$ in document $m$ \\
~~~~~~~~~~~~~sample $w\sim multinomial(\beta)$\\
\\\hline\hline
\end{tabular}
\caption{The generative model for running the LDA model on the collected white papers.}
\label{generative}
\end{figure}

\paragraph{LDA}
\begin{figure}
    \centering
        \includegraphics[width=3in]{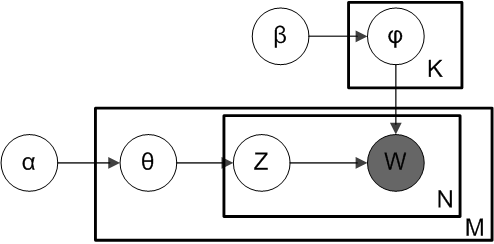}
        \caption{Overview of the LDA model. Suppose that we have $M$ white papers, $K$ topics and V different words.  Each white paper $m\in M$ is characterized by a document-topic distribution $\theta$, which is a $M\times K$ matrix. Each topic $k\in K$ is characterized by a topic-word distribution $\phi$, which is a $K\times V$ matrix.}
        \label{fig-LDA}
\end{figure}

We run a Latent Dirichlet Allocation (LDA) model \cite{blei2003latent} on the collected white papers. 
LDA is a generative statistical model that explains text documents with word clusters called ``topics" based on word-to-word co-occurrence.  
Each document is presented as a probabilistic distribution of latent topics,
 and each latent topic is characterized by a probabilistic distribution over words.
 The generative process of LDA is shown in Figure \ref{generative} and
 the figure of this process is illustrated  in 
  Figure \ref{fig-LDA}. 

\begin{table*}
\center
\begin{tabular}{llllll} \hline
topic1&topic2&topic3&topic4&topic5& topic6 \\
bet & doctor  & jesus  & cryptocurrency& purchase&game   \\
bets & medical& motherland& hash& consumer&casino\\
gambling & medicalchain& christ & cryptonote&goods&gaming\\
blackjack & health& buddhist&key  & wages&gamemaker\\
player &patients&john & crypto & prices&virtual\\
lounge & clinical & christians& security&finance&weapons  \\
dicing & alzheimer&mothership &private &deficit&play\\
poker & records & church   &scratchpad &pay&gift\\ 
seed &healthcare&mediation& charitable&transaction&equipment \\
gamble & diabetes& buddha    &signature  & cost&player \\  \hline
Ethbet\footnote{\url{https://ethbet.io/}}&  Medicalchain\footnote{\url{https://medicalchain.com/en/}} & Jesus\footnote{\url{https://jesuscoin.network/}}    &Tokenbox\footnote{\url{https://tokenbox.io/}} & Bitserial\footnote{\url{https://bitserial.io/}}& Funfair\footnote{\url{https://funfair.io/}}\\
Edgeless\footnote{\url{https://edgeless.io/}}& Ai-doctor\footnote{\url{http://aidoc.me/}}&  Lotos\footnote{\url{https://lotos.network/}}    &Biblepay&GoldReward\footnote{\url{https://goldreward.io/}}&Enjin\footnote{\url{}} \\ \hline\hline
\end{tabular}
\caption{Top words for different LDA topics, along with cryptocurrencies that are assigned to this topic. }
\label{LDA-topics}
\end{table*}

We use collapsed Gibbs sampling and run 100 iterations on the dataset. We show top words for different LDA topics in Table \ref{LDA-topics}, along with the white papers/cryptocurrencies 
that are assigned to that topic with the highest probability. 

We can see a clear semantic domain represented by each topic:
ICOs for gambling, games, medical care, religion, machine-networks,   cryptography, 
insurance, etc.

\subsection{Founding-Teams}
\label{Founding-Teams}
Out of 2,251 ICO projects, we were able to  collect the information of 1,230 founding teams. The following passage illustrates the type of descriptions used for founding teams in our dataset: 

\begin{quote}
{\em Justin Sun, born in 1990, master of University of Pennsylvania, bachelor of Peking University, founder and CEO of mobile social application Peiwo and TRON, the former chief representative for Greater China of Ripple. 2011 Asia Weekly Cover People; 2014 Davos Global Shaper; 2015 CNTV new figure of the year; 2017 Forbes Asia 30 under 30 entrepreneurs;2015/2017 Forbes China 30 Under-30s; The Only Millennial Student in the first batch of entrepreneurs at Hupan University, an elite business school established by Jack Ma, the founder of Alibaba Group. }
\end{quote}

\subsubsection{Extracting Features from Bios}
We aim to automatically extract the most important characteristics from bios of founding team members. 
We treat this as an NLP {\it tagging} problem \cite{toutanova2003feature,huang2015bidirectional,miller2004name,tjong2003introduction}.
Most tagging models are supervised or semi-supervised approaches, which require 
first designing annotation guidelines
(including selecting an appropriate set of types to annotate), then annotating a large corpus. 

We define 5 categories of tags: born-date, university, degree,  companies and awards received. 
We annotated 500 bios of different people and split the dataset into 0.8/0.1/0.1 for training, dev and testing.  
Features we employ include:

\begin{tightitemize}
\item 
Word level features:
\begin{tightitemize}
\item Unigrams, Bigrams 
\item Word window  size of 3
\item POS features
\item NER features
\end{tightitemize}
\item Letter level word features
\begin{tightitemize}
\item Starts with a capital letter?
\item Has all capital letters?
\item Has all lower case letters?
\item Has non initial capital letters?
\item Has numbers?
\item Is all numbers?
\item Letter prefixes and suffixes 
\end{tightitemize}
\item Dictionaries
\begin{tightitemize}
\item A dictionary of  world companies. We use Freebase API\footnote{\url{https://developers.google.com/freebase/v1/}} to augment the dataset by using synonyms. 
\item A dictionary of world universities. Freebase API is used as well. 
\end{tightitemize}
\end{tightitemize}
POS features 
 and NER features are pre-extracted using the Stanford NLP toolkit \cite{toutanova2003feature,manning2014stanford}. 

We use a CRF-based tagging model \cite{toutanova2003feature,huang2015bidirectional,lafferty2001conditional}. 
Accuracy for each category is shown in Table \ref{tagging-acc}. 
\begin{table}[t]
\center
\begin{tabular}{cc} \hline
born-date & 0.952 \\
university & 0.985 \\ 
degree & 0.892 \\
company &0.912 \\
award & 0.842 \\\hline
all & 0.947 \\\hline
\end{tabular}
\caption{Tagging accuracy for analyzing bios of ICO founding teams.}
\label{tagging-acc}
\end{table}

\section{\sysname: A supervised-learning model}
In this section, we detail the deep learning model for ICO rating. 
The model uses very little prior knowledge about ICO projects, but rather learns the importance of various features from the collected real-world dataset. 

\subsection{\sysname, a supervised-learning-based rating model}
The model we use here is a supervised learning model. 
In a standard supervised-learning setting, we wish to find a model $F$, that maps an input $x$ to an output $y$:
\begin{equation}
y = F (x)
\end{equation}

\paragraph{Input}
The input $x$ is an ICO project, which includes different aspects of its publicly available information. 
\paragraph{Output}  The output 
$y\in \{0,1\}$
is a binary variable indicating whether an ICO project is a scam.
We use a learning model to predict the proportional price change of any given currency  after a year of its ICO.
Out of the collected 2,251 projects,  we are able to collect the information of 1,482 projects with known price history and ICOs conducted at least one year before this work was done. 

At training time, we use the price change of an ICO project in a year as training signals, trying to predict this price change given its ICO information.
The predicting function $F$ is learned by maximizing the $L2$ distance between the predicted price change and the gold-standard price change:
\begin{equation}
\begin{aligned}
&c = \text{sigmoid}( \frac{\text{price}(t)}{\text{price(0)}} )\\
&\hat{c} = F(x) \\
&L = \sum_{(x, c)}  || \hat{c}- c ||^2 
\end{aligned}
\label{objective}
\end{equation}
where $t=\text{a year}$, and price(t) denotes the price of a currency after one year of its ICO.

At test time, we can use $F(x)$ to predict the price change, 
and 
 think a project is a scam if the predicted price is less than $m$ percent of its ICO price:

\begin{equation}
y=\left\{
\begin{aligned}
&1  \text{~~~~if~~} \frac{\text{price}(t)}{\text{price(0)}}\leq m \\
&0 \text{~~~~otherwise} \\
\end{aligned}
\right.
\end{equation}
$m$ can be set as requested. In this paper 
it is set to 0.01, 0.1 and 1.

We split the obtained ICO projects into training, dev, and testing in a 0.8/0.1/0.1 ratio.
For each value of $m$, 
an ICO project $x$ in the training dataset is paired with a gold-standard label $y$. $y$ is a binary value, indicating whether the project $x$
is a scam project. 
The value of $y$ can be different with respect to different split bar $m$. 

To note, it is inevitably  easier  to use the $L2$ distance between predicted prices and gold-standard prices for evaluation. 
But the value of this $L2$ distance looks more obscure to the  readers/investors. We thus transform the predicted price change to a more interpretable binary value $y$, in which case 
the evaluation can be transformed to whether a model is able to correctly predict a scam ICO project. 

\subsection{Features}
One of the key parts in supervised learning models is how to represent the input $x$. 
In this subsection, we will detail how we transform each aspect of an ICO (e.g., white paper, founding team, website) into a machine-readable vector $h_x$. 
\begin{figure*}
    \centering
        \includegraphics[width=5.5in]{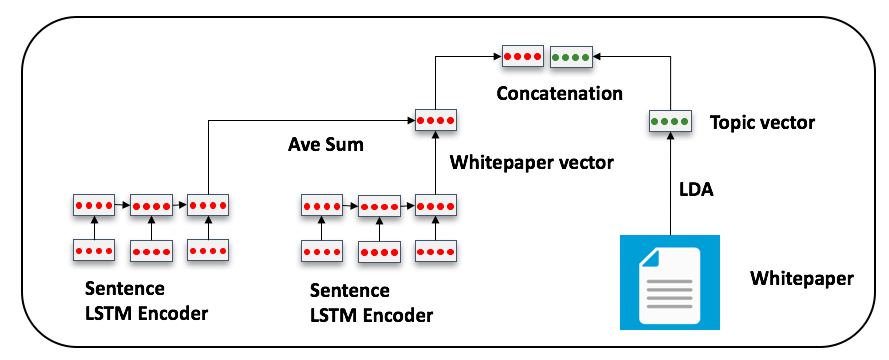}
        \caption{Transforming a white paper to a vector representation .}
        \label{Transforming-whitepaper}
\end{figure*}

\subsubsection{White papers}
We transform each white paper into a vector representation using deep learning methods. 
Each white paper $d$ consists a sequence of sentences $s$, $D=\{s_1,s_2,...,s_{n_D}\}$, where $n_D$ denotes the number of sentences
in the current white paper $s$. 
Each sentence $s$ consists of a sequence of words $w$, $s=\{w_1,w_2,...,w_{n_s}\}$, where $n_s$ denotes the number of words in sentence $s$. 
Each word $w$ is associated with a vector $e_w$. 

We adopt a hierarchical LSTM model \cite{serban2017hierarchical,li2015hierarchical} to map 
a white paper
$D$ to a vector representation $h_D$.
We first obtain representation vectors at the sentence level by
passing the vectors $e_w$ for each sentence's words
through four layers of LSTMs \cite{hochreiter1997long,gers1999learning}.
An LSTM associates each timestep with an input, memory and output gate, 
respectively denoted as $i_t$, $f_t$ and $o_t$.
For notational clarity, we distinguish $e$ and $h$, where $e_t$ denotes the vector for individual text units (e.g., words or sentences) at time step $t$ while $h_t$ denotes the vector computed by the LSTM model at time $t$ by combining $e_t$ and $h_{t-1}$. 
$\sigma$ denotes the sigmoid function. The vector representation $h_t$ for each time step $t$ is given by:

\begin{equation}
\Bigg[
\begin{array}{lr}
i_t\\
f_t\\
o_t\\
l_t\\
\end{array}
\Bigg]=
\Bigg[
\begin{array}{c}
\sigma\\
\sigma\\
\sigma\\
\text{tanh}\\
\end{array}
\Bigg]
W\cdot
\Bigg[
\begin{array}{c}
h_{t-1}\\
e_{t}\\
\end{array}
\Bigg]
\end{equation}
\begin{equation}
c_t=f_t\cdot c_{t-1}+i_t\cdot l_t\\
\end{equation}
\begin{equation}
h_{t}^s=o_t\cdot c_t
\end{equation}
where $W\in \mathbb{R}^{4K\times 2K}$
The vector output at the ending time-step is used to represent the entire sentence as 
\begin{equation}
e_s=h_{{end_s}}^w
\label{end}
\end{equation}

\begin{figure}
    \centering
        \includegraphics[width=2.6in]{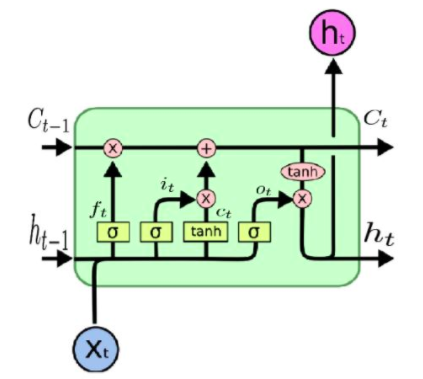}
        \caption{An illustration of a LSTM cell. Picture from Stratio RNN blog.\footnote{\url{http://www.stratio.com/blog/deep-learning-3-recurrent-neural-networks-lstm/}}}
        \label{fig-generative}
\end{figure}


The  representation $e_D$ for the current document/paragraph $D$ is obtained by sum over the presentation of 
all its containing sentences:
\begin{equation}
e_D  = \frac{1}{n_D}\sum_{s\in D} e_s 
\label{doc}
\end{equation} 

The best values for the LSTM parameters are unknown. 
The most straightforward way to learn the LSTM parameters is directly through the objective function in Eq. \ref{objective}. 
Unfortunately, training an LSTM model on a few thousand examples can easily lead to overfitting. 
We thus adopt an unsupervised approach, 
in which we train a 
skip-thought model 
 \cite{kiros2015skip,tang2017rethinking}.
 The skip-thought model is an encoder-decoder model based on sequence-to-sequence generation techniques \cite{sutskever2014sequence,luong2015effective,chung2014empirical}. 
The parameters of the LSTM are trained to maximize the predicted probability of 
each word in neighboring sentences given the current sentence. 
Our skip-thought model is exactly the same as \newcite{kiros2015skip} with the only difference being that word embeddings are initialized using
300-dimensional  Glove vectors \cite{pennington2014glove}. 
Given a pre-trained auto-encoder, we use the encoder to obtain sentence level representations using Eqs.\ 3, 4 and 5, then obtain document-level representations using 
Eq.\ \ref{doc}.

We also use the topic weights from the LDA model as white paper features. Let $K$ be the number of LDA topics, which is 50 in this work.
Each document can be represented as a vector describing a multinomial distribution over the 50 topics. 
We concatenate the deep learning vector and the LDA vector. The entire process is illustrated in Figure \ref{Transforming-whitepaper}.

\subsection{Founding Teams}
We map the founding team 
of an ICO project
to the following features:
\paragraph{Neural Network Features}
Again, we use a hierarchical neural model to map the bios of founding teams to a neural vector representation. 

\paragraph{Manually Designed Features} 
For each person, we obtained his or her full name, fed the name
into the Google+ API,\footnote{\url{https://developers.google.com/+/api/}}
and crawled the information. 
One key challenge here is that 
 one name can be mapped to multiple Google+ accounts.
 We assume a person's name is correctly mapped to a Google+ account if the company extracted from this person's bio can be found in their Google+ account.

Based on the information extracted (both from Section \ref{Founding-Teams} and crawled Google+ accounts), 
we include the following features:
\begin{tightitemize}
\item Whether founding team's bios can be found
\item Degree
\item Company the person previously works for
\item Age
\item Number of jobs the person had in the past 3 years before ICO
\item Whether the person is involved in other ICO projects
\end{tightitemize}

\subsection{Project Websites} 
We crawled the websites of each project and were able to retrieve information from 1,087 project websites. 
We map the website text to a vector representation using a hierarchical LSTM as described above. 
We also use a binary feature indicating whether an ICO project has a website.

\subsection{GitHub Accounts} 
We crawled the GitHub repository  (if it exists) of each ICO project.
A binary variable indicating whether an ICO project has a GitHub repository is included in the feature vector.
We handle GitHub README files similarly to the white papers, using an encoder-decoder model to map the file to a vector representation.
Additional features include the number of branches, the number of commits, the total lines of code, and the total number of files. 
We only consider the version before the time of ICO. 
\subsection{Other Features}
Additional features we consider include:
\begin{tightitemize}
\item Platform, e.g., ETH.
\item Total supply.
\item Unlimited or hard cap: an unlimited cap allows investors to send unlimited funding to the project’s ICO wallet. 
\end{tightitemize}

\subsection{Regression}
By concatenating white paper features, founding team features, website features and GitHub features,
each ICO project $x$ is associated with a vector $h_x$. 
The predicted price change $\hat{c}$
is given by:
\begin{equation}
\begin{aligned}
& \hat{c}= \frac{\exp (w^T h_x)}{1+\exp (w^T h_x)} \\
\end{aligned}
\end{equation} 
$w$ is the weight vector learned through
Eq. \ref{objective}
with stochastic gradient decent  \cite{bottou2010large} using with batch size set to 30. $L_2$ regularization is added. 
The number of iterations is treated as a hyperparameter to be tuned on the dev set. 

\section{Experiments}
We present experimental results and 
qualitative analysis in this section. 
\subsection{Results}
\begin{table}
\begin{tabular}{lccc}\hline
Feature type & Precision & Recall &F1 \\\hline
white paper & 0.26 & 0.92 & 0.41 \\\hline
GitHub & 0.24 & 0.93 & 0.38 \\\hline
founding team & 0.24 & 0.72 & 0.36 \\\hline
website & 0.22&0.75& 0.34 \\\hline
white paper + & \multirow{ 2}{*}{0.34} &\multirow{ 2}{*}{0.93} & \multirow{ 2}{*}{0.49} \\
GitHub& \\\hline
white paper + & \multirow{ 3}{*}{0.36} &\multirow{ 3}{*}{0.94} & \multirow{ 3}{*}{0.52} \\
GitHub +& \\
founding team &  \\\hline
white paper + & \multirow{ 5}{*}{0.37} &\multirow{ 5}{*}{0.95} & \multirow{ 5}{*}{0.53} \\
GitHub +& \\
founding team + &  \\
website +& \\
other features  &\\\hline
\end{tabular}
\caption{Results for identifying scam ICO projects with the scam bar $m$ is set to 0.01. The proportion of positive (scam) data points is $5.2\%$. }
\label{0-01}
\end{table}

\begin{table}
\begin{tabular}{lccc}\hline
Feature type & Precision & Recall &F1 \\\hline
white paper & 0.62 & 0.75 & 0.68 \\\hline
GitHub & 0.56 & 0.76 & 0.64 \\\hline
founding team & 0.41 & 0.72 & 0.52 \\\hline
website & 0.42&0.68& 0.52 \\\hline
white paper + & \multirow{ 2}{*}{0.70} &\multirow{ 2}{*}{0.80} & \multirow{ 2}{*}{0.75} \\
GitHub& \\\hline
white paper + & \multirow{ 3}{*}{0.72} &\multirow{ 3}{*}{0.82} & \multirow{ 3}{*}{0.76} \\
GitHub +& \\
founding team &  \\\hline
white paper + & \multirow{ 5}{*}{0.73} &\multirow{ 5}{*}{0.84} & \multirow{ 5}{*}{0.78} \\
GitHub +& \\
founding team + &  \\
website +& \\
other features  &\\\hline
\end{tabular}
\caption{Results for identifying scam ICO projects with the scam bar $m$ is set to 0.1. The proportion of positive (sam) data points is $28\%$.}
\label{0-1}
\end{table}

\begin{table}
\begin{tabular}{lccc}\hline
Feature type & Precision & Recall &F1 \\\hline
white paper & 0.72 & 0.69 & 0.70 \\\hline
GitHub & 0.74 & 0.68 & 0.71 \\\hline
founding team & 0.65 & 0.60 & 0.62 \\\hline
website & 0.63&0.59& 0.61 \\\hline
white paper + & \multirow{ 2}{*}{0.77} &\multirow{ 2}{*}{0.74} & \multirow{ 2}{*}{0.75} \\
GitHub& \\\hline
white paper + & \multirow{ 3}{*}{0.80} &\multirow{ 3}{*}{0.76} & \multirow{ 3}{*}{0.78} \\
GitHub +& \\
founding team &  \\\hline
white paper + & \multirow{ 5}{*}{0.83} &\multirow{ 5}{*}{0.77} & \multirow{ 5}{*}{0.80} \\
GitHub +& \\
founding team + &  \\
website +& \\
other features  &\\\hline
\end{tabular}
\caption{Results for identifying scam ICO projects with the scam bar $m$ is set to 1. The proportion of positive (sam) data points is $47\%$.}
\label{0-5}
\end{table}
Tables \ref{0-01}, \ref{0-1} and \ref{0-5}
present results for scam ICO project identification with respect to different values of  scam bar $m$. 
As the value of $m$ increases from 0.01 to 0.1, then to 1, the proportion of the scam projects increases, giving progressively higher precision and lower recall. 
The white paper and GitHub repository are the most important two classes of features,
 achieving approximate F1 scores of 0.7 when $m$ is set to 0.1 and 0.5.
By adding more features, we are able to get progressively better precision and recall. 
The model achieves 0.83  precision, 0.77 recall  and  0.80 F1 score 
 in predicting scam ICO projects in the $m=1$ setting, when all features are considered.

\subsection{Qualitative Analysis}
We need to rationalize the output from the model. 
Deep learning models are hard to be rationalized directly \cite{lei2016rationalizing,mahendran2015understanding,weinzaepfel2011reconstructing,vondrick2013hoggles}.
This is because neural network models operate like a black box:
using vector representations (as opposed to humaninterpretable
features) to represent  inputs, and
applying multiple layers of non-linear transformations.

Various  techniques have been proposed to make neural models interpretable \cite{koh2017understanding,montavon2017methods,mahendran2015understanding,li2015visualizing,li2016understanding}. The basic idea of these models is to build another learning or visualization model on top of a pre-trained neural model to for interpretation. 
We adopt  two widely used methods for neural model visualization purposes:

\paragraph{First-derivative Saliency} 
 has been widely used for visualizing and understanding neural models \cite{montavon2017methods,simonyan2013deep,li2015visualizing}.
The basic idea of this saliency method
is to compute 
the  contribution of each  cell/feature/representation to the final decision using the derivative of the final decision with respective to 
an input feature.

More formally, for a supervised  model, an
input $x$ is associated with a  class
label $y$. A  trained neural model associates
the pair (x, y) with a score $S_y(x)$. The saliency $w(e)$ of  any feature $e$ is given by:
\begin{equation}
w(e) =| \frac{\partial S_y(x)}{\partial(e)} |
\end{equation}
Suppose that one aspect (denoted by $a$, e.g., whitepaper) of an ICO project is represented by a vector $e\in$ vector($a$). For example, the vector for  
a whitepaper is the concatenation of its topic vector output from LDA and the neural vector output from the hierarchical LSTM model. 
The saliency of the aspect $a$ is thus the average sum of its containing vector units:
\begin{equation}
w(a) = \frac{1}{|\text{vector}(a)|} \sum_{e\in \text{vector}(a)}w(e) 
\end{equation}

\begin{figure}[!ht]
    \centering
        \includegraphics[width=3.3in]{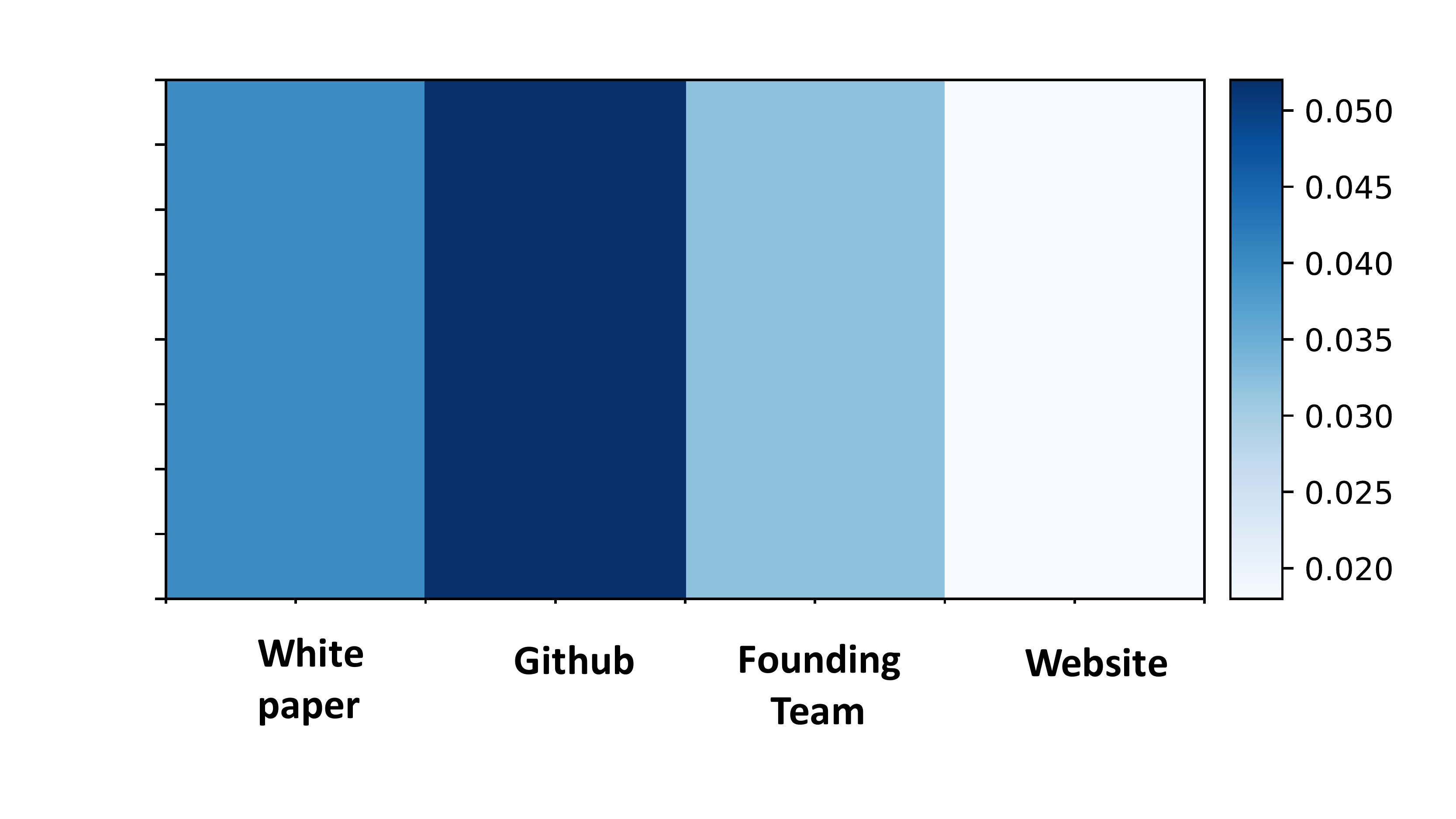}
        \caption{Heat map of whitepaper, github, founding team and website. }
        \label{heat-map}
\end{figure}

The saliency of different four aspects are illustrated in Figure \ref{heat-map}. As can be seen, whitepaper and github are the most salient aspects. 
 \paragraph{Feature Erasing} 
One problem with the first-derivative saliency method is that it is able to show which feature is important (in other words, salient),  but cannot tell how positively or negatively a particular feature contributes to a
decision. 
  We thus adopt the representation  erasing strategy for neural model visualization, which has been  used in a variety of previous work \cite{li2016understanding}.
The basic idea of this method is as follows:
how much a feature/cell/representation contributes to a decision is determined by the negative 
effect of erasing pieces of the representation. 

More formally, for a pre-trained supervised  model, 
$S(x, y=1)$ denotes the score that
the label of scam project
is assigned to 
 the input $x$.
$S(x \neg e, y=1)$ 
denotes the score that
the label of scam project
is assigned to 
 the input $x$ with feature $e$ being erased. 
If $e$ is a significant feature that leads the input ICO project $x$ to be a scam, 
$S(x, y=1)$ should be larger than $S(x \neg e, y=1)$. 
  Let $w(e, y=1)$ denote the influence of feature $e$ on class $y=1$ (a scam ICO project), which is given as follows: 
    \begin{equation}
w(e, y=1)  = \frac{S(x, y=1) - S(x \neg e, y=1) }{S(x \neg e, y=1)}
  \end{equation}

A negative value of $w(e, y=1)$ means feature $e$ positively contributes to the input $x$ being thought as a scam project. 
  By computing the derivative with respect to each LDA topic,
we are also able to rank LDA topics by the risking of being a scam. 
We manually labeled 10 LDA topics, each of which  has a clear meaning.
We compute the influence  score of each human-defined topic, as can be seen in Table \ref{LDA-score}.
ICOs on gaming, gambling and entertainment are more likely to scams than exchange, payment and smart contract. 

\begin{table}
\center
\begin{tabular}{cc}\hline
Topic &  Score \\\hline
Gaming & -1.62  \\
Gambling & -1.23 \\ 
Entertainment & -1.17 \\
 Religion &  -0.81 \\
 Medical Care &-0.23\\
Insurance & -0.12 \\
Payments &  -0.02 \\
Exchange & 0.14\\ 
Smart Contract & 0.22\\
Cryptography &0.28 \\\hline
\end{tabular}
\caption{Scam score for human-labeled LDA topic. Larger value means unless likely to be scam project.}
\label{LDA-score}
\end{table}

\section{Conclusion}
 ICOs have become one of the most controversial topics in the financial world.
For legitimate projects, they provide fairness  in crowdfunding, but
a lack of transparency, technical understanding and legality gives unscrupulous actors an incentive to launch scam ICO projects, bringing significant 
loss
to individual investors and making the world of cryptofinance fraught with danger.

In this paper, we proposed the first machine learning--based scam-ICO identification system. We find that a well designed neural network system is able to 
identify subtle warning signs hidden below the surface.
By integrating different types of information about the ICO, the system is able to predict whether the price of an cryptocurrency will go down. 
We hope the proposed system will help investors identify scam ICO projects and attract more  academic and public-sector work investigating this problem. 

\bibliographystyle{acl_natbib}
\bibliography{MMI_MT}
\end{document}